# AraDetox: A Multi-Dialect Arabic Detoxification Dataset

**Mo El-Haj**
VinUniversity, Vietnam
Lancaster University, UK
https://elhaj.uk

## Abstract

Arabic harmful-language detection has received considerable attention, yet Arabic text detoxification remains underexplored. We introduce AraDetox, a multi-dialect Arabic detoxification dataset comprising 10,500 harmful social-media posts and 84,000 detoxified rewrites generated using GPT-5 and Gemini 2.5 Flash across Modern Standard Arabic, Gulf, Levantine, and Egyptian Arabic. The generated outputs were assessed through human evaluation and automatic analyses of lexical change, semantic preservation, sentiment, and dialectal style. Results show that detoxification is primarily a meaning-preserving rewriting task: substantial lexical and structural reformulation is accompanied by consistently high semantic similarity. Human evaluation confirms successful harmful-language removal while largely preserving the original meaning. Dialectal analyses further indicate that the generated variants exhibit measurable stylistic alignment with reference Arabic dialect corpora. Comparison with existing resources highlights two complementary approaches to detoxification: minimal-edit lexical substitution and meaning-preserving reformulation. Our findings demonstrate that large-scale Arabic detoxification resources can be constructed through LLM-assisted generation and human verification. The dataset is publicly available at https://github.com/ArabicNLP-UK/AraDetox to support future research on Arabic detoxification, safe text generation, and multi-dialect Arabic NLP.

## 1 Introduction

Research on harmful language in Arabic has largely focused on detection, resulting in numerous datasets and models for hate speech, offensive language, and toxicity classification (Al Mandhari et al., 2024). Comparatively little attention has been paid to *text detoxification*, where harmful content is rewritten into a safer form while preserving its meaning. Creating detoxification datasets is particularly challenging because it requires annotators to rewrite offensive, politically sensitive, and often dialectal content while preserving meaning, stance, target, and communicative intent. As a result, manual dataset creation is both costly and time-consuming (Logacheva et al., 2022; Dementieva et al., 2024). Recent advances in large language models (LLMs) provide a promising alternative, enabling large-scale generation of fluent rewrites that remove harmful language while retaining the underlying message. Unlike harmful-language detection, detoxification requires balancing two competing objectives: removing harmful expressions while preserving the original claim, stance, target, and communicative intent. This challenge is particularly pronounced in Arabic social-media discourse, where criticism, sarcasm, political disagreement, and identity-related language are often intertwined with offensive expressions. As a result, successful detoxification cannot be reduced to simple lexical substitution, but instead requires broader meaning-preserving reformulation.

We introduce AraDetox, a large-scale multi-dialect Arabic detoxification dataset containing 10,500 harmful social-media posts and 84,000 detoxified rewrites generated using GPT-5 and Gemini 2.5 Flash across four Arabic varieties: Modern Standard Arabic (MSA), Gulf Arabic, Levantine Arabic, and Egyptian Arabic. The dataset was created using an LLM-assisted synthesis pipeline followed by human quality control by native Arabic speakers. We evaluate AraDetox using lexical change, semantic similarity, sentiment analysis, dialectal style analysis, and human assessment. Our findings show that Arabic detoxification is best characterised as a meaning-preserving rewriting task. Across models and dialects, substantial lexical and structural reformulation is accompanied by consistently high semantic similarity. Human evaluation further confirms that harmful language can be removed while largely preserving the original meaning. Beyond introducing a new resource, our work demonstrates that existing Arabic harmful-language datasets can be transformed into large-scale multi-dialect detoxification resources through LLM-assisted generation and human verification. The dataset is publicly available through the AraDetox repository[1]. The main contributions of this work are threefold. First, we introduce a large-scale Arabic detoxification resource containing eight rewrites for each source post across two LLM families and four Arabic varieties. Second, we provide a multi-perspective evaluation covering lexical reformulation, semantic preservation, sentiment, dialectal style, and independent human assessment. Third, we release the dataset to support reproducible research on Arabic detoxification and dialect-aware safe generation.

## 2 Related Work

Research on harmful language in Arabic has focused primarily on detection, resulting in datasets and models for hate speech, offensive language, and abusive-language classification (Mubarak et al., 2021, 2022; Alghamdi et al., 2024), as well as Arabic language models such as AraBERT (Antoun et al., 2020) and ARBERT/MARBERT (Abdul-Mageed et al., 2021). Comparatively little attention has been paid to *text detoxification*, which aims to rewrite harmful content into safer forms while preserving meaning. ParaDetox introduced a parallel-data setting based on toxic and non-toxic paraphrase pairs (Logacheva et al., 2022), while subsequent work extended detoxification to multilingual settings through MultiParaDetox (Dementieva et al., 2024, 2025). These resources established detoxification as a generation task distinct from harmful-language detection, but existing Arabic resources remain relatively limited in scale and provide only a single detoxified rewrite per source text. For Arabic, the original MultiParaDetox dataset was derived from approximately 2,100 candidate posts and resulted in 1,181 manually annotated source–detoxified pairs after filtering and quality control. For our empirical comparison, we use the publicly available Arabic evaluation split released through Hugging Face[2], which contains 400 source–detoxified pairs. Recent work has also shown that LLMs can generate high-quality synthetic datasets and annotations, making them increasingly useful for data creation and augmentation (Gilardi et al., 2023; Long et al., 2024; Nadăș et al., 2025; Yong et al., 2024). AraDetox builds on these developments by providing detoxified rewrites for Arabic social-media content across Modern Standard Arabic, Gulf, Levantine, and Egyptian Arabic using GPT-5 and Gemini. Unlike existing Arabic detoxification resources, AraDetox emphasises semantic preservation while retaining the original intent, stance, and target of the source text. AraDetox substantially expands the scale of Arabic detoxification resources and, to the best of our knowledge, is the first Arabic detoxification dataset to provide rewrites generated across multiple dialects and multiple LLMs within a single framework.

[1] https://github.com/ArabicNLP-UK/AraDetox

[2] https://huggingface.co/datasets/textdetox/multilingual_paradetox

## 3 Methodology

### 3.1 Dataset Construction

*AraDetox* is derived from the Arabic Hate Speech Superset (Tonneau et al., 2024), which integrates ten Arabic hate-speech and abusive-language datasets: T-HSAB (Haddad et al., 2019), JHSC (Ahmad et al., 2024), MLMA (Ousidhoum et al., 2019), L-HSAB (Mulki et al., 2019), OSACT (Mubarak et al., 2020), Let-Mi (Mulki and Ghanem, 2021), Saudi Tweets (Alshaalan and Al-Khalifa, 2020), AraCOVID19-MFH (Ameur and Aliane, 2021), Brothers (Albadi et al., 2018), and Alsafari et al. (Alsafari et al., 2020). From the 16,000 harmful posts available in the superset, we selected 10,500 posts while retaining coverage across all ten constituent datasets. No class balancing was applied, and the source datasets were not equally represented because they varied substantially in size. The selection was intended to provide broad coverage of the available harmful-language data rather than a balanced classification sample. Table 1 summarises the resulting resource.

Using GPT-5 and Gemini 2.5 Flash, we generated detoxified rewrites in MSA, Gulf, Levantine, and Egyptian Arabic. Models were instructed to preserve the original meaning, target, stance, sentiment, and communicative intent while removing profanity, insults, slurs, dehumanising language, discriminatory expressions, and threats. For the dialectal variants, outputs were additionally required to adapt the source text into Gulf, Levantine, or Egyptian Arabic. These outputs therefore combine two transformations: harmful-language detoxification and adaptation to the requested Arabic variety.

| Statistic | Value |
|---|---|
| Source posts | 10,500 |
| Detoxified outputs | 84,000 |
| Arabic varieties | MSA, Gulf, Levantine, Egyptian |
| Generation models | GPT-5, Gemini 2.5 Flash |
| Outputs per source post | 8 |
| Outputs generated by GPT-5 | 42,000 |
| Outputs generated by Gemini | 42,000 |
| Total words | 1,318,257 |
| Average words per output | 15.69 |
| Median words per output | 13 |
| Human-evaluation sample | 300 source posts |
| Human-evaluated outputs | 2,400 |
| Annotators | 3 |
| Annotation records | 7,200 |

Table 1: Summary statistics for AraDetox.

The MSA outputs primarily represent detoxification within the same language variety, whereas the Gulf, Levantine, and Egyptian outputs jointly involve detoxification and dialect adaptation. Consequently, lexical and structural differences between the source posts and the dialectal outputs cannot be attributed exclusively to detoxification. We therefore evaluate harmful-language removal, meaning preservation, and dialectal style as separate dimensions and interpret the dialectal results as evidence of joint detoxification and dialect adaptation. This produced eight detoxified variants per source post, yielding 84,000 detoxified instances and more than 1.3 million words. The full prompts are provided in Appendix A. Dataset construction incorporated automatic validation, regeneration of invalid outputs, and human quality control. Automatic checks identified missing fields, malformed JSON, null values, duplicated outputs, incomplete generations, and source–output alignment errors. Safety refusals, generic moderation responses, summaries, external-observer descriptions, and outputs that substantially altered or added to the source meaning were treated as invalid and regenerated. A native Arabic-speaking reviewer subsequently verified that the regenerated outputs performed the requested detoxification task, preserved the source meaning, target, stance, and communicative intent, and reflected the requested Arabic variety. Further details of the generation procedure and quality-control workflow are provided in Appendix B.

As an additional quality-assurance step, 300 source posts were randomly selected from the 10,500-post collection. The sample was stratified to preserve the distribution of the ten source datasets across the ten source datasets and available harm categories. For each selected post, all eight detoxified variants were evaluated by three native Arabic-speaking annotators, yielding 2,400 generated outputs and 7,200 annotation records. The annotators involved in this evaluation were distinct from the native Arabic speaker who performed quality control during dataset construction. Appendix E reports the distribution of the full dataset and the human-evaluation sample across the constituent source datasets.

### 3.2 Evaluation Design

We evaluate AraDetox from five complementary perspectives: lexical change, semantic preservation, sentiment shift, dialectal style, and human evaluation. We additionally compare AraDetox with MultiParaDetox-Ar (Dementieva et al., 2024, 2025) to contextualise its detoxification behaviour relative to an existing Arabic detoxification resource. A key challenge in detoxification is preserving meaning while removing harmful content. Offensive expressions often convey stance, criticism, and emotion rather than functioning as removable noise. For example, the insult لا انت الكلب يا خنزير ("No, you are the dog, you pig") can be rewritten as لا، بل إنك أنت من يتصرف بطريقة غير مقبولة ("No, rather, you are the one behaving in an unacceptable manner"), preserving the criticism while removing the offensive language. Appendix F presents additional qualitative examples from AraDetox. F1 score is not used as a generation-quality metric because detoxification does not have a single correct reference sequence from which token-level true positives, false positives, and false negatives can be meaningfully derived. Multiple lexically different rewrites may all be valid. We therefore combine surface-change measures, semantic similarity, sentiment analysis, dialectal-style analysis, and human judgements, with each evaluation component measuring a distinct requirement of successful detoxification.

#### 3.2.1 Lexical Change

We measure lexical change between each source post and its detoxified counterpart using six surface-level metrics: lexical overlap, Jaccard similarity, character-level edit distance, percentage of changed characters, word-level edit distance, and output length. Lexical overlap and Jaccard similarity measure how much of the original vocabulary is retained, with higher values indicating more conservative rewriting. Character- and word-level edit distances quantify the amount of rewriting required to transform the source post into the detoxified out-put, while the percentage of changed characters normalises this change by source length. Output length is included to capture whether detoxification tends to shorten, preserve, or expand the original text. These measures allow us to distinguish minimal-edit detoxification from broader sentence-level reformulation.

### 3.2.2 Semantic Preservation

To quantify semantic fidelity, we compute cosine similarity between each source post and its detoxified counterpart using multilingual-e5-large (Wang et al., 2024). This embedding-based measure assesses whether the detoxified text remains semantically close to the original despite lexical and structural changes. To complement the pairwise similarity analysis, we examine the global organisation of the embedding space using UMAP. By visualising source posts, detoxified outputs, and external neutral Arabic corpora in a shared semantic space, we assess whether detoxified texts remain close to their source content while exhibiting characteristics associated with naturally occurring neutral language.

### 3.2.3 Sentiment and Neutrality Analysis

Detoxification should remove harmful language without eliminating criticism, disagreement, or negative opinions. We therefore examine whether detoxification systematically shifts sentiment by increasing neutral or positive predictions at the expense of negative ones. This follows the broader use of sentiment analysis for examining evaluative language and affect across different domains and languages, including financial discourse, social media, and multilingual settings (El-Haj et al., 2016; Alwakid et al., 2022; Hunter et al., 2023; Huy et al., 2026). Sentiment distributions are computed using three CAMeLBERT sentiment classifiers: CAMeLBERT-DA, CAMeLBERT-MSA, and CAMeLBERT-Mix (Inoue et al., 2021). Changes in positive, neutral, and negative predictions are analysed before and after detoxification to assess whether rewriting reduces hostile affect while preserving the original stance.

### 3.2.4 Dialectal Adaptation

AraDetox includes detoxified variants in Gulf, Levantine, and Egyptian Arabic. To examine whether these rewrites exhibit stylistic alignment with their intended varieties, we compare them against reference dialect corpora (Zaidan and Callison-Burch, 2014) using TF–IDF cosine similarity over word and character n-grams. This analysis provides corpus-level evidence of lexical and orthographic alignment, but it is sensitive to topic, register, corpus composition, and vocabulary shared across Arabic varieties. It is therefore treated as an exploratory measure of dialectal style rather than definitive validation of dialect authenticity. For example, the MSA phrase أغلقوا الشارع ("they blocked the street") may be realised as سادين الشارع in Gulf Arabic, سكروا الشارع in Levantine Arabic, and قفلوا الشارع in Egyp-tian Arabic, illustrating meaning-preserving adaptation across Arabic varieties.

### 3.2.5 Human Evaluation Protocol

To assess the quality of the generated outputs, three native Arabic-speaking annotators independently reviewed a random sample of 300 source posts. For each post, all eight detoxified variants were evaluated, yielding 2,400 generated outputs and 7,200 annotation records. Each output was assessed for offence removal, meaning preservation, and introduction of new content. Offence removal was annotated using a three-way scheme: offence not removed, offence removed, or not applicable when the source text did not contain offensive content. Meaning preservation and new content were annotated using binary judgements. The full annotation guidelines are provided in Appendix D. Dialect authenticity was not included as a separate human-evaluation criterion. The dialectal findings therefore rely on corpus-level stylistic analysis and should not be interpreted as direct human validation of native-like dialect use.

### 3.2.6 Comparison with Existing Resources

To contextualise AraDetox, we compare it descriptively with MultiParaDetox-Ar using the same lexical, semantic, and sentiment-based measures. The resources differ in source-domain distribution, size, annotation procedure, and the initial sentiment and severity of their source texts. The comparison is therefore intended to characterise their rewriting behaviour rather than establish that one resource or construction method is superior. It also contrasts the relatively minimal-edit strategy represented by MultiParaDetox-Ar with the broader meaning-preserving reformulation and dialect adaptation represented by AraDetox.

## 4 Results

### 4.1 Lexical Change

Table 2 summarises the lexical transformations introduced by all eight detoxification systems. Overlap and Jaccard measure vocabulary retention between the original and detoxified texts, while CharEdit, Char%, WordEdit, and Word% capture character- and word-level modifications. Across all variants, detoxification results in substantial modification of the original text, indicating that the generated outputs frequently involve sentence-level reformulation rather than simple replacement of offensive expressions. Among all systems, GPT-MSA is the most conservative. It achieves the lowest character-level edit distance (42.19), the lowest proportion of modified characters (61.58%), the lowest word-level edit distance (11.77), the lowest word-level modification rate (94.15%), and produces the shortest outputs (11.28 words on average). GPT-MSA therefore preserves more of the original surface form than the other systems.

In contrast, Gemini-MSA performs more extensive rewriting than GPT-MSA, exhibiting larger character- and word-level modifications, higher proportions of changed characters, and longer outputs. The dialectal variants generally involve even greater reformulation. Across both models, output lengths increase from an average of 13.21 words in the original posts to 15.66–16.79 words, while character-level modification rates exceed 78%. Normalised word-level edit rates range from 115.79% to 122.94%. Values above 100% occur because word-level edit distance is normalised by the original text length and includes insertions, deletions, and substitutions, allowing the number of edit operations to exceed the number of words in the source text.

GPT-Gulf exhibits the highest lexical overlap (0.2942) and Jaccard similarity (0.1719), indicating greater retention of source vocabulary. In contrast, Gemini-MSA and Gemini-Egyptian show the lowest overlap scores, suggesting a greater degree of lexical reformulation. The lexical analyses suggest that Arabic detoxification is best characterised as a meaning-preserving rewriting task rather than simple lexical substitution. Whether these extensive modifications preserve the underlying message is examined through the semantic and human evaluations that follow.

### 4.2 Semantic Preservation

Table 3 reports semantic similarity between the original posts and the AraDetox variants using multilingual E5 embeddings. Across all variants, semantic similarity remains consistently high, with mean scores ranging from 0.9234 to 0.9308 and all examples exceeding the 0.70 similarity threshold. These results suggest that the generated rewrites preserve the core meaning of the source posts despite the substantial lexical modifications reported in Section 4.1.

The dialect-aware variants achieve slightly higher similarity scores than the MSA variants, with Levantine obtaining the highest mean similarity (0.9308), followed by Egyptian and Gulf Arabic. GPT-5 and Gemini also exhibit comparable levels of semantic preservation, with mean similarities of 0.9234 and 0.9249, respectively. The relatively small differences across variants suggest that semantic fidelity is largely unaffected by the choice of dialect or generation model.

The lexical and semantic results show that substantial reformulation does not necessarily come at the expense of semantic fidelity. The generated outputs often differ considerably from the source posts at the surface level, while remaining close in meaning. This supports treating detoxification as a meaning-preserving rewriting task rather than simple lexical substitution.

Figures 1 and 2 visualise the embedding space using UMAP. Across all variants, substantial overlap is observed between the original posts and their detoxified counterparts, with no clear separation between source texts and generated rewrites. Similar patterns are observed for the dialect-aware variants, whose outputs occupy largely the same regions of the embedding space as the original posts. These visualisations are consistent with the high semantic-similarity scores reported in Table 3 and provide additional evidence that detoxification and dialect adaptation preserve the semantic structure of the source content.

### 4.3 Sentiment Shift

We investigate whether detoxification influences negative affect using three Arabic sentiment models: CAMeLBERT-DA, CAMeLBERT-MSA, and CAMeLBERT-Mix. Sentiment is not treated as a direct measure of toxicity, since a post may remain critical, negative, or emotionally charged without being abusive, hateful, or discriminatory. Rather, sentiment provides an additional perspective on how detoxification alters the tone of the text while preserving its meaning and intent.

| System | Overlap | Jaccard | CharEdit | Char% | WordEdit | Word% | Len |
|---|---|---|---|---|---|---|---|
| GPT-MSA | 0.2062 | 0.1339 | 42.19 | 61.58 | 11.77 | 94.15 | 11.28 |
| GPT-Gulf | 0.2942 | 0.1719 | 45.52 | 81.24 | 13.08 | 115.79 | 15.99 |
| GPT-Levantine | 0.2737 | 0.1610 | 45.90 | 80.66 | 13.24 | 116.14 | 15.66 |
| GPT-Egyptian | 0.2537 | 0.1454 | 48.07 | 84.72 | 13.67 | 120.51 | 16.04 |
| Gemini-MSA | 0.2007 | 0.1201 | 52.74 | 83.61 | 14.59 | 119.38 | 16.76 |
| Gemini-Gulf | 0.2342 | 0.1353 | 49.90 | 83.90 | 14.39 | 122.94 | 16.79 |
| Gemini-Levantine | 0.2440 | 0.1471 | 47.69 | 78.88 | 13.94 | 117.08 | 16.27 |
| Gemini-Egyptian | 0.2188 | 0.1263 | 49.93 | 81.26 | 14.50 | 120.85 | 16.75 |

Table 2: Lexical change statistics.

| Variant | Mean | Median | Min. | >0.90 |
|---|---|---|---|---|
| **Levantine** | **0.9308** | **0.9349** | **0.7883** | **80.31%** |
| Egyptian | 0.9274 | 0.9307 | 0.7789 | 77.93% |
| Gulf | 0.9256 | 0.9293 | 0.7700 | 76.15% |
| Gemini | 0.9249 | 0.9278 | 0.7691 | 73.30% |
| GPT-5 | 0.9234 | 0.9278 | 0.7504 | 71.05% |

Table 3: Semantic similarity using multilingual E5 embeddings.

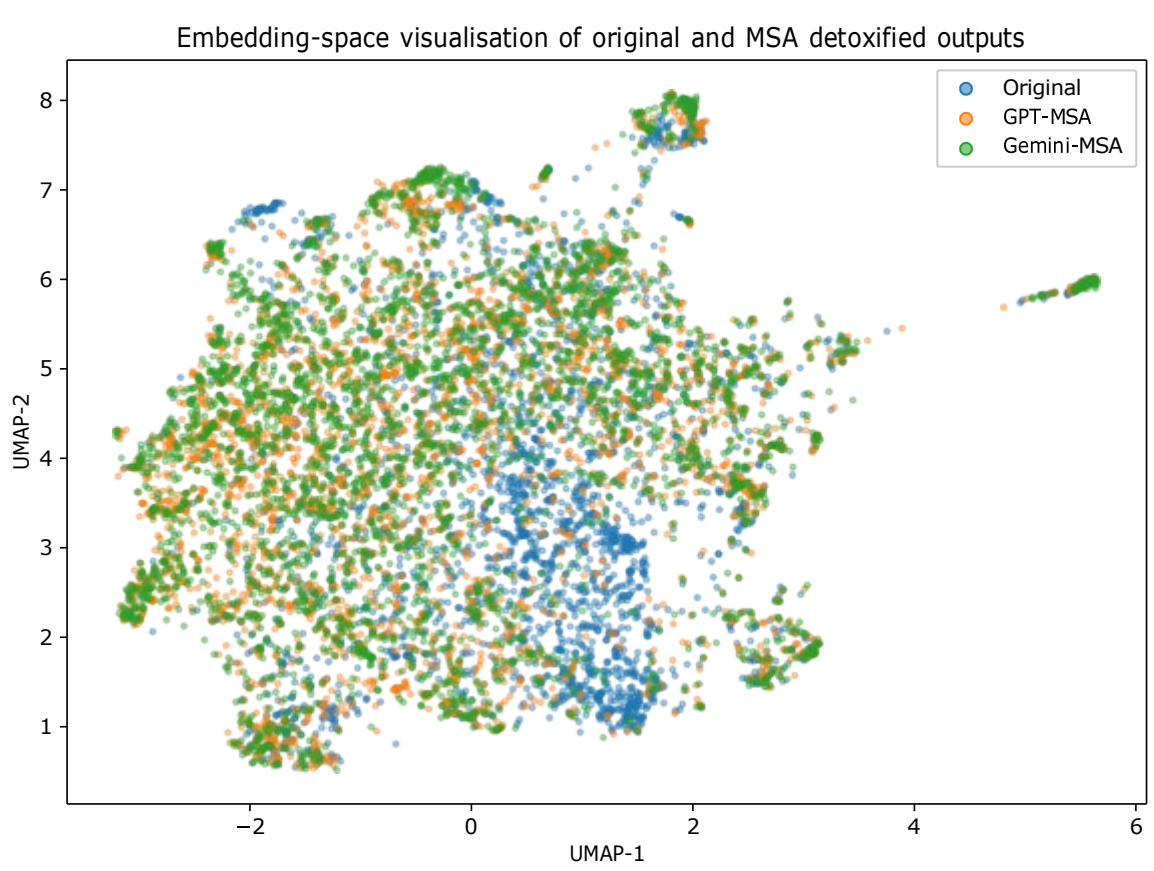

Figure 1: UMAP visualisation of original posts, GPT-MSA detoxifications, and Gemini-MSA detoxifications using multilingual E5 embeddings.

Table 4 and Figure 3 report the percentage of posts classified as negative before and after detoxification. The original posts are classified as negative at high rates across all three sentiment models, ranging from 72.41% to 75.47%. After detoxification, the strongest reduction is observed for the MSA variants. GPT-MSA reduces negative sentiment to 49.20%–53.77%, while Gemini-MSA reduces it to 53.19%–60.26%. MSA detoxification substantially softens the affective tone of harmful posts.

The dialectal variants also reduce negative sentiment, but to a more limited extent. Across both GPT and Gemini, Gulf, Levantine, and Egyptian outputs remain closer to the original sentiment distribution, with negative rates generally between 65% and 74%. Dialectal rewriting preserves more of the original emotional tone, even after harmful expressions have been removed. The pattern is also visible in Figure 4, where the MSA variants show the clearest drop in average negative sentiment, while the dialectal variants remain nearer to the original posts.

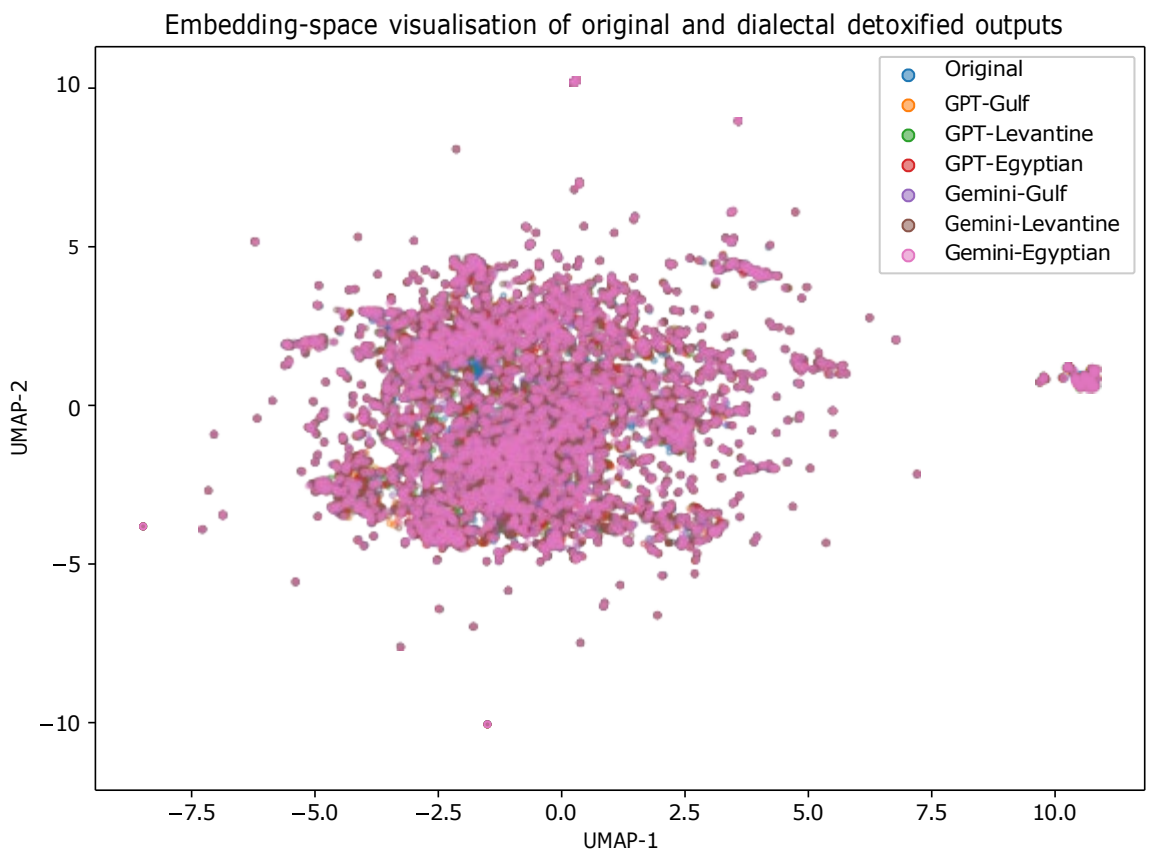

Figure 2: UMAP visualisation of original posts and GPT/Gemini dialectal detoxified variants using multilingual E5 embeddings.

| Model | Orig. | GPT-MSA | GPT-Gulf | GPT-Lev. | GPT-Egy. |
|---|---|---|---|---|---|
| DA | 74.29 | 53.77 | 70.32 | 71.14 | 69.38 |
| MSA | 75.47 | 53.48 | 71.23 | 73.67 | 70.44 |
| Mix | 72.41 | **49.20** | 68.62 | 69.56 | 66.36 |

| Model | Orig. | Gem-MSA | Gem-Gulf | Gem-Lev. | Gem-Egy. |
|---|---|---|---|---|---|
| DA | 74.29 | 55.79 | 68.07 | 70.13 | 70.07 |
| MSA | 75.47 | 60.26 | 69.69 | 72.24 | 70.21 |
| Mix | 72.41 | 53.19 | 65.45 | 67.42 | 65.18 |

Table 4: Negative sentiment before and after detoxification using CAMeLBERT-DA, MSA and Mix models.

These results indicate that part of the negative affect detected by sentiment models is associated with abusive wording, insults, profanity, and other harmful linguistic constructions. Once such language is removed, the resulting texts are generally perceived as less negative, particularly in MSA. At the same time, the dialectal variants show that detoxification does not necessarily eliminate criticism, disagreement, or emotional stance. Rather, the generated

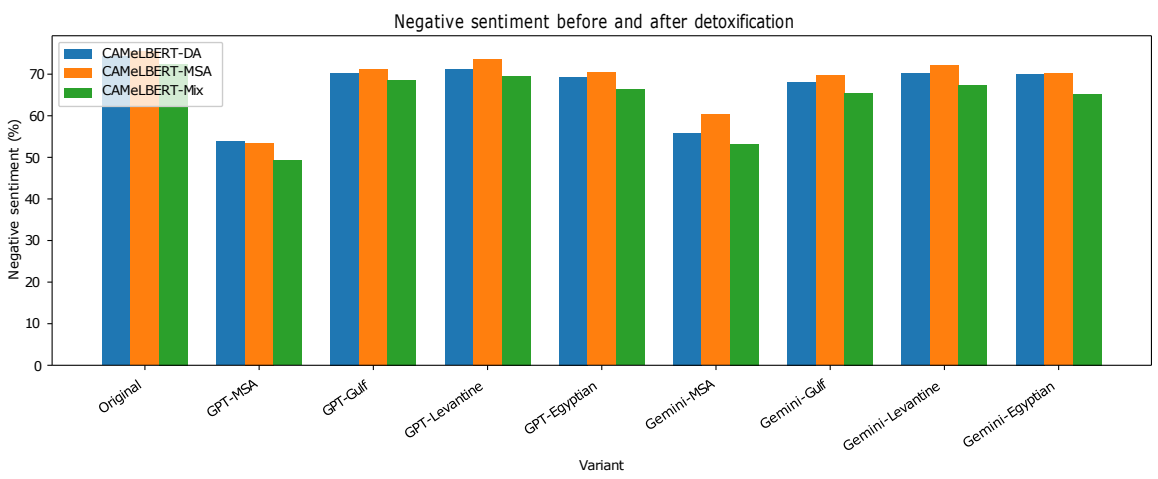

Figure 3: Negative sentiment across GPT and Gemini variants using three CAMeLBERT models.

outputs preserve much of the communicative intent while reducing harmful expression.

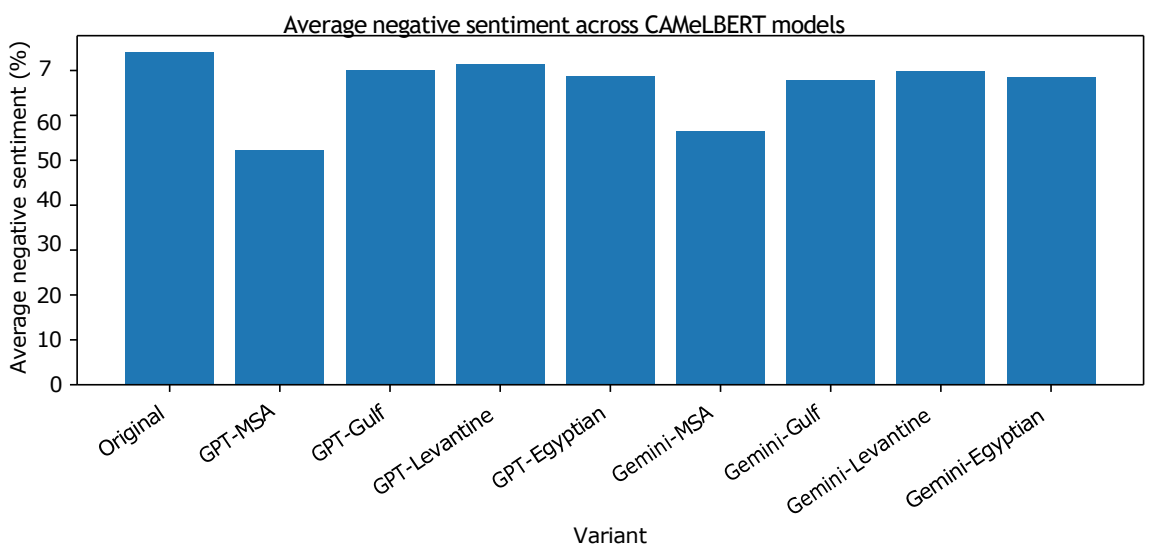

Figure 4: Average negative-sentiment rate across the three CAMeLBERT models.

## 4.4 Dialectal Style Analysis

AraDetox includes detoxified rewrites in MSA, Gulf, Levantine, and Egyptian Arabic. To examine corpus-level dialectal style, we compare the generated variants with reference texts from the Arabic-Dialects corpus using TF–IDF cosine similarity over word and character n-grams.

Figures 5 and 6 show the resulting similarity matrices. The word n-gram results provide the clearest evidence of dialectal alignment. Both MSA systems align most strongly with the MSA reference corpus, with Gemini-MSA achieving the highest MSA similarity (0.373), followed by GPT-MSA (0.350). The Egyptian variants likewise show strong alignment with the Egyptian reference corpus, with GPT-Egyptian obtaining the highest similarity (0.406), followed by Gemini-Egyptian (0.383).

The Gulf and Levantine results are less distinct. Gemini-Gulf aligns most strongly with the Gulf reference corpus (0.285), and Gemini-Levantine aligns most strongly with the Levantine reference corpus (0.297), suggesting clearer dialect-specific stylistic patterns. In contrast, GPT-Gulf and GPT-Levantine show their highest similarity with the Egyptian reference corpus, likely reflecting lexical overlap among informal Arabic varieties and the strong signal of the Egyptian reference set. The character n-gram analysis broadly confirms the MSA and Egyptian findings but provides weaker separation between Gulf and Levantine. Character-level features capture orthographic and morphological properties shared across Arabic dialects, making them less discriminative than word n-grams.

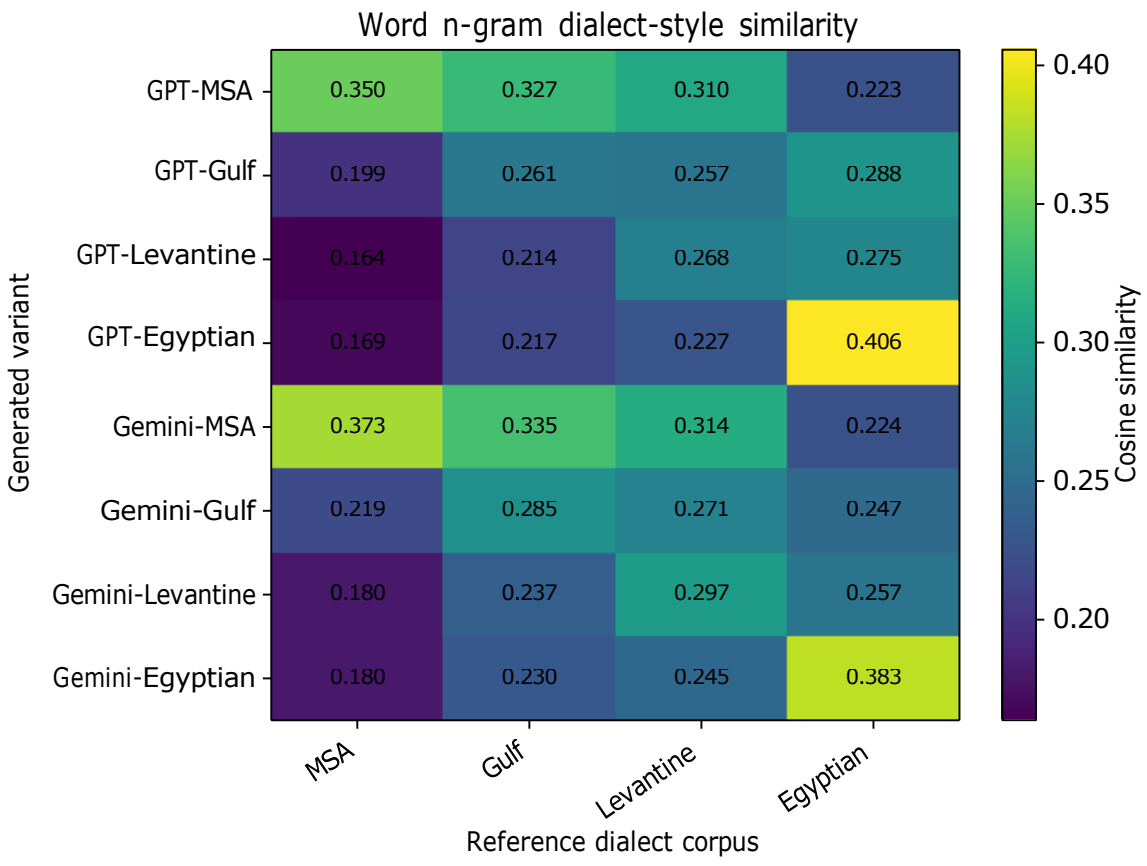

Figure 5: Word n-gram similarity to reference dialect corpora.

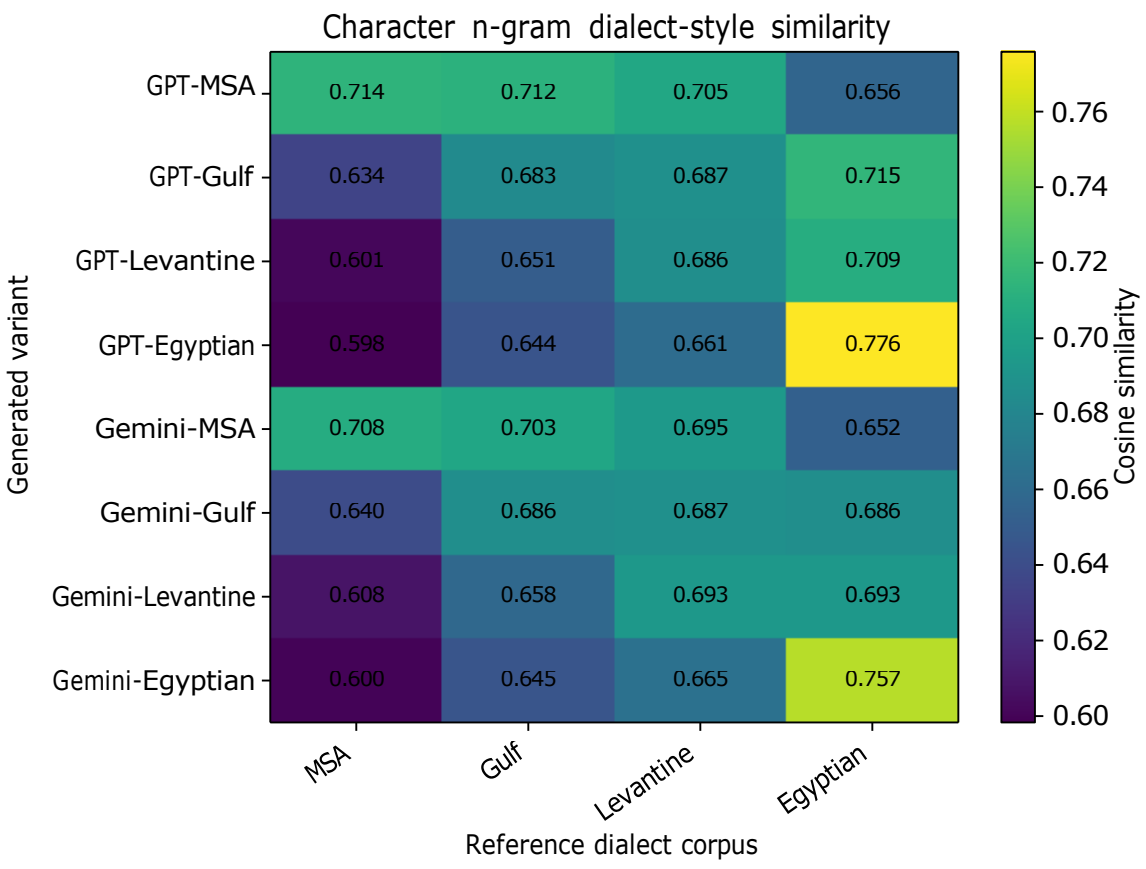

Figure 6: Character n-gram similarity to reference dialect corpora.

The results provide measurable evidence of corpus-level dialectal style, particularly for MSA and Egyptian Arabic. Gemini also shows stronger intended-variety alignment for Gulf and Levantine than GPT-5. However, the reference corpora consist largely of general social-media discussions and forum-style interactions, whereas AraDetox is derived from harmful, highly negative, politically sensitive, and identity-related discourse. Topic, register, and corpus composition may therefore influence the reported similarities. The weaker distinction between Gulf and Levantine further reflects the substantial lexical and orthographic overlap between informal Arabic varieties. These results should consequently be interpreted as evidence of stylistic alignment rather than definitive evidence of native-like dialect authenticity.

### 4.5 Human Evaluation

Automatic metrics provide useful insights into lexical change, semantic similarity, and sentiment shifts, but they cannot directly determine whether harmful content has been removed while preserving the intended meaning of the original post. To address this limitation, we conducted a human evaluation following the protocol described in Section 3.2.5. Three native Arabic-speaking annotators independently evaluated all outputs in the evaluation sample. The annotators represented Saudi Gulf Arabic, Jordanian Levantine Arabic, and Egyptian Arabic and were regular users of Arabic social media. They were independent of the LLM-generation process and distinct from the native Arabic speaker who performed quality control during dataset construction.

The evaluation focused on three criteria central to detoxification: *offence removal*, *meaning preservation*, and *introduction of new content*. These dimensions assess whether generated rewrites successfully remove harmful language, retain the original communicative intent, and avoid introducing information not present in the source text. To minimise potential bias, annotators were not informed that the texts had been generated by LLMs.

Inter-annotator agreement was substantial across all evaluation dimensions. Fleiss' $\kappa$ values ranged from 0.715 to 0.795, indicating consistent judgements among annotators and supporting the reliability of the evaluation results (Table 5).

| Criterion | Agreement (%) | Fleiss' $\kappa$ |
|---|---|---|
| Offence Removal | 96.44 | 0.795 |
| Meaning Preservation | 95.30 | 0.715 |
| New Content | 92.00 | 0.716 |

Table 5: Inter-annotator agreement for human evaluation.

| Variant | Offence | Meaning | New |
|---|---|---|---|
| Gemini MSA | **92.90** | **92.57** | **12.42** |
| Gemini Gulf | 92.35 | 91.46 | 15.52 |
| Gemini Levantine | 92.24 | 91.69 | 13.75 |
| Gemini Egyptian | 91.91 | 89.36 | 17.52 |
| GPT MSA | 87.25 | 86.25 | 13.08 |
| GPT Gulf | 87.25 | 91.57 | 24.83 |
| GPT Levantine | 87.47 | 91.80 | 24.28 |
| GPT Egyptian | 87.58 | 91.69 | 24.17 |

Table 6: Human evaluation results (%). Higher is better except for New Content.

Table 6 shows strong performance across all three evaluation dimensions. Offence-removal rates range from 87.25% to 92.90%, showing that harmful language can generally be removed without substantially altering the underlying message. Gemini achieves the highest offence-removal scores across all language varieties.

Meaning-preservation rates are also high, ranging from 86.25% to 92.57%. The generated rewrites therefore largely retain the original claims, targets, stances, and communicative intent of the source posts despite modifications introduced during detoxification.

The clearest differences between the model families emerge in the introduction of new content. Gemini introduces additional information in 12.42%–17.52% of cases, whereas GPT's dialectal variants do so in 24.17%–24.83% of outputs. Although GPT tends to produce more expansive rewrites, its meaning-preservation scores remain comparable to those of Gemini, suggesting that the additional content generally supplements rather than alters the core meaning of the source text.

The human evaluation findings are consistent with the automatic analyses. Both model families are effective at removing harmful language while preserving the intended meaning of the original posts. The results also point to different rewriting strategies: Gemini tends to produce more conservative reformulations, whereas GPT more frequently elaborates on the source content while maintaining similar levels of meaning preservation.

### 4.6 Comparison with Existing Arabic Detoxification Resources

To better understand the characteristics of AraDetox, we compare it descriptively with MultiParaDetox-Ar (Dementieva et al., 2024, 2025), one of the few publicly available Arabic detoxification resources. The comparison covers lexical change, semantic similarity, and sentiment shift. Because the two resources differ substantially in source distribution, size, annotation procedure, and source-text negativity, the results should be interpreted as a comparison of dataset characteristics rather than as a controlled benchmark.

Table 7 reveals clear differences in rewriting behaviour. MultiParaDetox-Ar follows a relatively conservative editing strategy, exhibiting higher word overlap (0.559), lower character-level modification (32.96%), and lower word-level modification (46.72%). In contrast, the AraDetox variants exhibit substantially greater reformulation, with word-overlap scores ranging from 0.201 to 0.294 and con-siderably higher character- and word-level modification rates than those observed in MultiParaDetox-Ar. Despite these larger lexical and structural changes, semantic similarity remains consistently high across all AraDetox variants. MultiParaDetox-Ar achieves the highest E5 similarity (0.957), consistent with its minimal-edit approach. The AraDetox variants obtain E5 similarities between 0.919 and 0.930, indicating that extensive rewriting can occur while remaining semantically close to the original content. Among the AraDetox variants, the dialect-aware rewrites achieve the strongest semantic preservation, with Gulf and Levantine Arabic producing the highest similarity scores. These findings suggest that the two resources operationalise detoxification through different rewriting strategies. MultiParaDetox-Ar primarily reflects *minimal-edit detoxification*, characterised by conservative lexical modifications and high surface-level similarity to the source text. In contrast, AraDetox emphasises *meaning-preserving reformulation*, retaining the original claim, target, and stance while permitting substantially broader lexical and structural rewriting. This distinction is particularly evident in the dialectal variants, where detoxification is combined with adaptation to dialect-specific linguistic conventions.

| Dataset | Pairs | Overlap | Char. % | Word % | E5 |
|---|---|---|---|---|---|
| MultiParaDetox-Ar | 400 | **0.559** | **32.96** | **46.72** | **0.957** |
| AraDetox GPT-MSA | 10,500 | 0.206 | 61.58 | 94.15 | 0.924 |
| AraDetox GPT-Gulf | 10,500 | 0.294 | 81.24 | 115.79 | 0.930 |
| AraDetox GPT-Lev. | 10,500 | 0.274 | 80.66 | 116.14 | 0.930 |
| AraDetox GPT-Egy. | 10,500 | 0.254 | 84.72 | 120.51 | 0.927 |
| AraDetox Gemini-MSA | 10,500 | 0.201 | 83.61 | 119.38 | 0.919 |
| AraDetox Gemini-Gulf | 10,500 | 0.234 | 83.90 | 122.94 | 0.924 |
| AraDetox Gemini-Lev. | 10,500 | 0.244 | 78.88 | 117.08 | 0.928 |
| AraDetox Gemini-Egy. | 10,500 | 0.219 | 81.26 | 120.85 | 0.925 |

Table 7: Comparison of MultiParaDetox-Ar and AraDetox.

We further compare the datasets using negative-sentiment predictions from three CAMeLBERT sentiment models. Table 8 should be interpreted with caution because the source datasets differ substantially. Only around 10–11% of MultiParaDetox-Ar source texts are classified as negative, compared with 72–75% for AraDetox, reflecting the more strongly negative nature of the harmful-language corpora from which AraDetox was derived.

Even with these differences, distinct patterns emerge. MultiParaDetox-Ar shows relatively small reductions in negative sentiment (1.22–2.39 percentage points), whereas AraDetox MSA variants produce much larger reductions. GPT-MSA reduces negative predictions by 20.51–23.21 percentage points across the three sentiment models, while Gemini-MSA achieves reductions of 15.21–19.22 points. The dialectal variants show smaller reductions, typically between 1.80 and 7.23 points. These results reinforce the distinction between detoxification and sentiment modification. AraDetox is not designed to convert negative opinions into positive ones, but to express criticism, disagreement, or opposition without insults, slurs, threats, or abusive language. As a result, many detoxified outputs remain negative in sentiment while no longer containing harmful expressions.

| Model | Dataset | Orig. | Detox | Red. |
|---|---|---|---|---|
| DA | MultiParaDetox-Ar | 9.92 | 7.53 | 2.39 |
| | GPT-MSA | 74.29 | 53.77 | 20.51 |
| | GPT-Gulf | 74.29 | 70.32 | 3.96 |
| | GPT-Lev. | 74.29 | 71.14 | 3.14 |
| | GPT-Egy. | 74.29 | 69.38 | 4.90 |
| | Gemini-MSA | 74.29 | 55.79 | 18.50 |
| | Gemini-Gulf | 74.29 | 68.07 | 6.22 |
| | Gemini-Lev. | 74.29 | 70.13 | 4.15 |
| | Gemini-Egy. | 74.29 | 70.07 | 4.22 |
| MSA | MultiParaDetox-Ar | 11.00 | 9.78 | 1.22 |
| | GPT-MSA | 75.47 | 53.48 | 21.99 |
| | GPT-Gulf | 75.47 | 71.23 | 4.24 |
| | GPT-Lev. | 75.47 | 73.67 | 1.80 |
| | GPT-Egy. | 75.47 | 70.44 | 5.03 |
| | Gemini-MSA | 75.47 | 60.26 | 15.21 |
| | Gemini-Gulf | 75.47 | 69.69 | 5.78 |
| | Gemini-Lev. | 75.47 | 72.24 | 3.23 |
| | Gemini-Egy. | 75.47 | 70.21 | 5.26 |
| Mix | MultiParaDetox-Ar | 10.14 | 7.78 | 2.36 |
| | GPT-MSA | 72.41 | 49.20 | 23.21 |
| | GPT-Gulf | 72.41 | 68.62 | 3.79 |
| | GPT-Lev. | 72.41 | 69.56 | 2.85 |
| | GPT-Egy. | 72.41 | 66.36 | 6.05 |
| | Gemini-MSA | 72.41 | 53.19 | 19.22 |
| | Gemini-Gulf | 72.41 | 65.45 | 6.96 |
| | Gemini-Lev. | 72.41 | 67.42 | 4.99 |
| | Gemini-Egy. | 72.41 | 65.18 | 7.23 |

Table 8: Negative sentiment before and after detoxification using CAMeLBERT-DA, MSA and Mix.

## 5 Conclusion

We introduce *AraDetox*, a large-scale Arabic detoxification dataset containing 10,500 harmful social-media posts and 84,000 detoxified rewrites across MSA, Gulf, Levantine, and Egyptian Arabic. The resource combines LLM-assisted generation with automatic validation, native-speaker quality control, and independent human evaluation. The findings show that Arabic detoxification frequently requires substantial reformulation while preserving the source claim, target, stance, and communicative intent. Human evaluation indicates high rates of offence removal and meaning preservation, although some outputs introduce additional content. The dialectal variants also show measurable corpus-level alignment with their intended Arabic varieties. AraDetox provides a substantial resource for developing and evaluating Arabic detoxification systems, while highlighting the need for continued work on dialect validation, pragmatic meaning preservation, and downstream evaluation.

## 6 Limitations

This study has several limitations. First, AraDetox was constructed primarily through GPT-5 and Gemini 2.5 Flash rather than large-scale human rewriting. Although automatic validation and native-speaker quality control were applied, the outputs may retain model-specific stylistic patterns and should not be treated as naturally occurring human rewrites.

Second, detailed human evaluation covered 300 source posts, corresponding to 2,400 generated outputs, and therefore represents only a subset of the full 84,000-output resource. Rare source types or harm categories may be underrepresented.

Third, semantic-preservation analyses rely partly on multilingual embedding models. High cosine similarity does not guarantee preservation of pragmatic meaning, sarcasm, humour, presupposition, target, or subtle stance. Human judgements partly address this limitation but cannot eliminate it.

Fourth, dialectal evaluation relies on corpus-level word and character n-gram similarity, which may be influenced by topic, register, corpus composition, and shared vocabulary across Arabic varieties. As dialect authenticity was not separately evaluated by human annotators, these results should be interpreted as evidence of stylistic alignment rather than native-like dialect use.

Fifth, comparison with MultiParaDetox-Ar is descriptive because the resources differ in size, source distributions, and annotation procedures. We also do not yet assess whether models trained or fine-tuned on AraDetox outperform those trained on existing Arabic detoxification datasets. A controlled downstream benchmark remains future work.

Finally, the dataset was generated using two proprietary frontier model families. Evaluating smaller open-weight models would help determine sensitivity to model scale, architecture, and training data. AraDetox is also derived from Arabic social-media content and may not generalise to long-form, news, or formal political discourse.

The AraDetox dataset is publicly available at https://github.com/ArabicNLP-UK/AraDetox to support transparency and reproducibility.

## 7 Ethical Considerations

This work involves harmful Arabic social-media content, including offensive, abusive, political, sectarian, and identity-related language. The purpose of the study is to develop safer Arabic NLP resources and evaluation methods for neutralising harmful language. We minimise the reproduction of harmful examples in the paper and report results in aggregate form.

The study included a human evaluation component involving up to ten native Arabic-speaking adult annotators. Ethical approval was granted by an Institutional Ethical Review Board for Biomedical Research in Vietnam. Annotators provided informed consent and were given detailed annotation guidance prior to participation. No personal, sensitive, or health-related data were collected, and all procedures were conducted in accordance with institutional policy, Vietnamese regulations, and internationally recognised ethical standards.

The detoxified outputs should not be treated as authoritative or universally acceptable rewrites, since judgements of harmfulness and acceptable neutralisation are shaped by social, cultural, dialectal, and political context. We also recognise that automatic detoxification can introduce risks, including over-sanitising legitimate political expression, weakening the speaker's intended stance, or removing important evidence of abuse. Any practical deployment of Arabic detoxification systems should therefore include human oversight and clear moderation guidelines.

Outputs that produced refusals, malformed responses, summaries, external-observer descriptions, or substantial meaning changes were treated as invalid and regenerated during dataset construction. The released resource distinguishes the complete LLM-generated collection from the independently human-evaluated subset. Human-evaluation labels for offence removal, meaning preservation, and new-content introduction are provided for the evaluated outputs, allowing users to identify and filter unsuccessful or uncertain generations. The complete synthetic collection should not be treated as a gold-standard dataset of human-authored detoxifications.

# Appendix

## A Detoxification Prompt

For each source post, GPT-5 and Gemini 2.5 Flash were instructed to generate detoxified rewrites in Modern Standard Arabic (MSA), Gulf Arabic, Levantine Arabic, and Egyptian Arabic. The prompt emphasised meaning preservation while removing harmful language.

> Rewrite the following Arabic social-media post into four civil and non-offensive Arabic versions:
>
> 1. Modern Standard Arabic (MSA)
> 2. Gulf Arabic
> 3. Levantine Arabic
> 4. Egyptian Arabic
>
> Preserve the original meaning, target, stance, criticism, sentiment, and communicative intent.
>
> Remove profanity, insults, slurs, dehumanising language, discriminatory expressions, threats, calls for violence, and unnecessarily inflammatory wording.
>
> Do not defend, refute, explain, summarise, fact-check, or respond to the statement. Rewrite directly from the perspective of the original author.
>
> Produce natural social-media language in the requested variety and return the output as JSON:

```
{
  "msa_detox": "...",
  "gulf_detox": "...",
  "levantine_detox": "...",
  "egyptian_detox": "..."
}
```

## B Generation Procedure and Quality Control

Detoxified rewrites were generated using GPT-5 and Gemini 2.5 Flash through their respective APIs. Generation was performed in batches using structured JSON output to ensure consistent formatting across all generated variants. The final resource contains 42,000 GPT-generated rewrites and 42,000 Gemini-generated rewrites, corresponding to eight detoxified variants for each of the 10,500 source posts. Additional generations were performed during prompt development, quality control, and regeneration of invalid outputs. GPT-5 and Gemini 2.5 Flash were selected as representatives of two contemporary frontier LLM families with strong multilingual generation capabilities. Gemini 2.5 Flash provides a favourable trade-off between generation quality, latency, and cost for large-scale dataset construction, while GPT-5 was included to provide a complementary model family and enable cross-model comparisons. Our objective was not to compare LLM performance, but to investigate whether meaning-preserving detoxification remained consistent across different model families and Arabic language varieties.

| Parameter | GPT-5 | Gemini 2.5 Flash |
|---|---|---|
| API access | OpenAI API | Gemini API |
| Temperature | API default | API default |
| Top-$p$ | API default | API default |
| Maximum output tokens | API default | API default |
| Structured JSON output | Yes | Yes |
| Prompt development | Iterative pilot testing | Iterative pilot testing |
| Automatic regeneration | Yes | Yes |
| Human quality control | Yes | Yes |

Table 9: Generation settings used during dataset construction. Parameters not explicitly specified in the API requests used the provider's default values.

**Prompt development.** The final prompt was developed through iterative pilot testing on 100 source posts. Earlier prompt versions were revised to reduce safety refusals, summaries, external-observer responses, malformed JSON, unsupported additions, and changes to the original target or stance. The final prompt was selected because it produced the most consistent structured outputs while preserving the source perspective and communicative intent. Table 9 summarises the generation settings used for both models.

The generation scripts implemented automatic validation checks to detect missing fields, malformed JSON, duplicated outputs, null values, and incomplete generations. Outputs that failed these checks were automatically flagged for regeneration. Retry mechanisms were also implemented to handle API failures, empty responses, malformed outputs, and safety-related refusals. Following generation, each batch underwent a quality-control stage performed by a native Arabic speaker. The objective of this stage was not to rewrite or annotate the outputs, but to verify that the generation process had been completed correctly and that the resulting texts were suitable for inclusion in the dataset.

Quality control focused on several aspects. The reviewer checked that each generated output corresponded to the correct source post and preserved its intended meaning, target, stance, and communicative intent. They also verified that the MSA, Gulf, Levantine, and Egyptian variants reflected the intended language variety and exhibited natural linguistic usage. The reviewer inspected the outputs for technical issues, including alignment errors between source posts and generated rewrites, inconsistencies across dialectal variants, and generation failures that were not detected automatically.

A particular challenge arose from the safety mechanisms of contemporary LLMs. When presented with highly offensive or sensitive content, models occasionally refused the task and produced safety-oriented responses rather than detoxified rewrites (e.g. *I cannot fulfil this request. I am programmed to be a helpful and harmless AI assistant*). In other cases, the models adopted an external observer perspective by describing the content of a post rather than rewriting it from the perspective of the original author (e.g. *The speaker appears unhappy with the government's stance*). Such outputs were considered invalid because they failed to perform the intended detoxification task.

When these issues occurred, the affected instances were regenerated and, where necessary, additional clarification was provided that the task formed part of a research experiment focused on meaning-preserving detoxification. The reviewer also checked for hallucinated content, omissions of important information, and cases where the generated text substantially altered the meaning of the original post.

This quality-control stage served as a human verification process rather than an annotation exercise. The reviewer did not manually rewrite the posts. Their role was to ensure that the LLMs correctly executed the detoxification task and that the generated outputs were complete, aligned with the source posts, linguistically appropriate, and suitable for subsequent analysis.

## C Experimental Pipeline

The experimental pipeline used in this paper consists of the following stages:

1. Collect harmful Arabic social-media posts from the Arabic Hate Speech Superset.
2. Generate detoxified Modern Standard Arabic (MSA) rewrites using GPT-5 and Gemini 2.5 Flash.
3. Generate detoxified Gulf Arabic rewrites using GPT-5 and Gemini 2.5 Flash.
4. Generate detoxified Levantine Arabic rewrites using GPT-5 and Gemini 2.5 Flash.
5. Generate detoxified Egyptian Arabic rewrites using GPT-5 and Gemini 2.5 Flash.
6. Compute lexical-change metrics between orig-

inal and detoxified texts.
7. Compute semantic similarity using multilingual E5 embeddings.
8. Visualise embedding spaces using UMAP.
9. Measure sentiment shift using three CAMeL-BERT sentiment models.
10. Evaluate dialectal style similarity using authentic Arabic dialect corpora.
11. Conduct human evaluation using three native Arabic-speaking annotators.
12. Compute inter-rater agreement and aggregate evaluation statistics.

## D Human Annotation Guidelines

Human annotators assessed each generated output according to the following criteria:

1. **Offence Removal**: whether harmful language present in the source text was successfully removed.
2. **Meaning Preservation**: whether the main meaning and communicative intent of the source text were preserved.
3. **New Content**: whether the generated output introduced information not present in the source text.

Offence removal was annotated using a three-way scheme:

- 0 = offence not removed,
- 1 = offence removed,
- 2 = not applicable (no offence present in the source text).

All remaining criteria were annotated using binary judgements, where 1 indicates that the criterion was satisfied and 0 indicates that it was not.

## E Human-Evaluation Sample Distribution

Table 10 compares the distribution of the complete AraDetox source collection with that of the 300-post human-evaluation sample.

| Source dataset | Evaluation sample | Evaluation sample (%) |
|---|---|---|
| JHSC | 30 | 10.00 |
| MLMA | 30 | 10.00 |
| T-HSAB | 30 | 10.00 |
| L-HSAB | 30 | 10.00 |
| AraCOVID19-MFH | 30 | 10.00 |
| Brothers | 30 | 10.00 |
| Saudi Tweets | 30 | 10.00 |
| Alsafari et al. | 30 | 10.00 |
| Let-Mi | 30 | 10.00 |
| OSACT | 30 | 10.00 |
| **Total** | **300** | **100.00** |

Table 10: Distribution of the 300 source posts included in the human-evaluation sample, with 30 posts sampled from each dataset.

## F Qualitative Examples

Tables 11 and 12 present representative examples from AraDetox. The examples illustrate a range of detoxification behaviours, including insult mitigation, meaning-preserving reformulation, stance preservation, and dialect-specific adaptation. The examples also highlight differences between GPT-5 and Gemini, with some outputs favouring direct lexical substitution and others employing broader paraphrastic rewriting.

| Source | GPT MSA | GPT Gulf | GPT Levantine | GPT Egyptian |
|---|---|---|---|---|
| انت انسانه تافهه | أنتِ شخصية سطحية | تصرفك سطحي ومخيّب للآمال | تصرفك سطحي ومخيّب للآمال | تصرفك سطحي ومخيّب للآمال |
| الملاحده بقر ياخي | أرى أن بعض الملحدين يفتقرون إلى المنطق في طرحهم لهذه القضايا | الملاحدة تفكيرهم سطحي يا أخي، ونقاشهم دايم يلف ويدور | الملحدين تفكيرهم سطحي يا زلمة، ونقاشهم دايم بلف وبدور | الملحدين تفكيرهم سطحي يا جدع، وكلامهم كتير بيلف ويدور |
| حمير | أشخاص يفتقرون إلى الحد الأدنى من الوعي والفهم | تصرفاتكم غير مقبولة ومحبطة | تصرفاتكم فعلاً مستفزة ومخيبة | تصرفاتكم دي مستفزة ومخيبة |
| كلي خرا مع محبتي | مع كامل الاحترام، لا أوافقك الرأي | مع محبتي، تصرفك مستفز وما كان له داعي | مع محبتي، تصرفك مستفز وما الو داعي | مع محبتي، تصرفك مستفز وملوش أي لزمة |

Table 11: Representative detoxification examples from AraDetox across GPT-5 variants. The examples illustrate meaning-preserving detoxification and dialect-specific adaptation across MSA, Gulf, Levantine, and Egyptian Arabic.

| Source | Gemini MSA | Gemini Gulf | Gemini Levantine | Gemini Egyptian |
|---|---|---|---|---|
| انت انسانه تافهه | أرى أن مقاربتكِ للأمور تفتقر إلى الجديّة والعمق | أشوف إن طريقتكِ في التعامل مع الأمور ما فيها جدية أو عمق | أنا بشوف إنه طريقتكِ بالتعامل مع الأمور ما فيها جدية أو عمق | أنا شايفة إن طريقتكِ في التعامل مع الأمور بتفتقر للجدية والعمق |
| الملاحده بقر ياخي | إن الملحدين يتسمون بقصور في التفكير | الملاحده عندهم قصور في فهم بعض الأمور ياخي | الملاحده تفكيرهم سطحي يا أخي | يا أخي، الواحد بيستغرب أوي من طريقة تفكير بعض الملحدين |
| حمير | قصور واضح في الفهم والإدراك | للأسف، يبدو أن البعض يجد صعوبة في استيعاب الأمور الواضحة | يا جماعة، في تصرفات بتدل على قلة تفكير واضحة | مفيش أي منطق في اللي بيحصل ده |
| كلي خرا مع محبتي | لا تضيع وقتك في هذا الهراء، مع خالص تقديري | ما عليك أمر، بلا هالحكي الفاضي، مع خالص تقديري | خلصنا من هالحكي الفاضي، مع كل احترامي | بلا كلام فارغ، مع كل احترامي |

Table 12: Representative detoxification examples from AraDetox across Gemini variants. The examples illustrate meaning-preserving detoxification and dialect-specific adaptation across MSA, Gulf, Levantine, and Egyptian Arabic.